\documentclass[letterpaper, 10 pt, conference]{ieeeconf}  % Comment this line out if you need a4paper

\IEEEoverridecommandlockouts                              % This command is only needed if 
\usepackage{amsmath}
\usepackage{amssymb}
\usepackage{mathtools}
\usepackage{booktabs}
\usepackage{subcaption}  % For creating subfigures
\usepackage{caption} 
\usepackage{color}
\usepackage[pdftex]{graphicx}
\usepackage{amsfonts} 
\usepackage{booktabs}%
\usepackage{array}
\usepackage{caption}
\usepackage{subcaption}
\usepackage{comment}
\usepackage[pdftex]{graphicx}
\usepackage{dblfloatfix}
\usepackage{comment}
\usepackage{cite}
\usepackage{amsfonts} 
\usepackage[skip=3pt,font=scriptsize]{caption}
\usepackage{amsmath}
\usepackage{algorithm}
\usepackage{algpseudocode}
\usepackage{comment}
\title{\LARGE \bf Reinforcement Learning Driven Multi-Robot Exploration via Explicit Communication and Density-Based Frontier Search}

\author{Gabriele Calzolari, Vidya Sumathy, Christoforos Kanellakis, George Nikolakopoulos% <-this % stops a space
\thanks{This work was partially supported by the Wallenberg AI, Autonomous Systems and Software Program (WASP) funded by the Knut and Alice Wallenberg Foundation, and by the European Union's Horizon Europe Research and Innovation Program, under the Grant Agreement No. 101119774 SPEAR.}
\thanks{The authors are within the Robotics and AI Group, Department of Computer Science, Electrical and Space Engineering, Luleå University of Technology, Sweden. Corresponding author's e-mail: gabcal@ltu.se}}

\begin{document}

\maketitle
\thispagestyle{empty}
\pagestyle{empty}

%%%%%%%%%%%%%%%%%%%%%%%%%%%%%%%%%%%%%%%%%%%%%%%%%%%%%%%%%%%%%%%%%%%%%%%%%%%%%%%%
\begin{abstract}
Collaborative multi-agent exploration of unknown environments is crucial for search and rescue operations. Effective real-world deployment must address challenges such as limited inter-agent communication and static and dynamic obstacles. This paper introduces a novel decentralized collaborative framework based on Reinforcement Learning to enhance multi-agent exploration in unknown environments. Our approach enables agents to decide their next action using an agent-centered field-of-view occupancy grid, and features extracted from $\text{A}^*$ algorithm-based trajectories to frontiers in the reconstructed global map. Furthermore, we propose a constrained communication scheme that enables agents to share their environmental knowledge efficiently, minimizing exploration redundancy. The decentralized nature of our framework ensures that each agent operates autonomously, while contributing to a collective exploration mission. Extensive simulations in Gymnasium and real-world experiments demonstrate the robustness and effectiveness of our system, while all the results highlight the benefits of combining autonomous exploration with inter-agent map sharing, advancing the development of scalable and resilient robotic exploration systems.
\end{abstract}

%%%%%%%%%%%%%%%%%%%%%%%%%%%%%%%%%%%%%%%%%%%%%%%%%%%%%%%%%%%%%%%%%%%%%%%%%%%%%%%%
\section{INTRODUCTION}

Decentralized collaborative reinforcement learning for multi-agent exploration has emerged as a paramount research field in robotics due to its potential to enhance the robustness and autonomy of the task execution in complex scenarios. Unlike traditional centralized architectures, these approaches allow autonomous agents to independently navigate and explore their surroundings, leveraging collaborative strategies to overcome the limitations of consistently sharing inter-agent knowledge with a centralized unit. This capability is especially crucial for the exploration of large-scale, real-world applications where constraints bound transmission of data between the robots, such as in search and rescue operations \cite{drew2021multi}, environmental monitoring \cite{zheng2020multi}, and autonomous exploration \cite{tan2023deep}. However, decentralized systems face challenges such as ensuring effective inter-agent communication, and managing partial observability of the environment.

This research extends the architecture proposed in \cite{marl_exp} for decentralized and collaborative multi-agent exploration of unknown environments using reinforcement learning. In particular, the exploring agents can autonomously make decisions about whether to share exploration data with networked robots or the navigation goals they should reach. To support the integration of physical hardware, these decisions are guided by information readily accessible from the robotic systems, such as an agent-centered field-of-view (FOV) occupancy grid, features extracted from the agent's reconstruction of the environment, and the configuration of the communication network. To address real-world exploration challenges and utility in search and rescue scenarios, the proposed research focuses on paramount aspects such as inter-agent proximity-based communication and the presence of both static and dynamic obstacles. The exploration task aims for at least one agent to develop a sufficiently extensive map of the environment tracking static obstacles. Therefore, the primary contributions are:

\begin{itemize}
    \item A decentralized collaborative architecture for autonomous multi-agent exploration based on reinforcement learning with constrained inter-agent data transmission encompassed in the agent's action space. The overall architecture is designed to leverage data from real robotic platforms.
    \item A novel feature-extraction methodology based on A*-estimated trajectories leading to frontiers in the agent's occupancy global grid map, for density-based frontier exploration, and a compact per-agent representation of the other transmitting agents within the same communication network.
    \item Experimental validation of the proposed framework in a custom-built exploration arena in the lab, using multiple TurtleBot3 mobile robots.
\end{itemize}

\begin{figure*}[t]
\centering
\includegraphics[width=\linewidth]{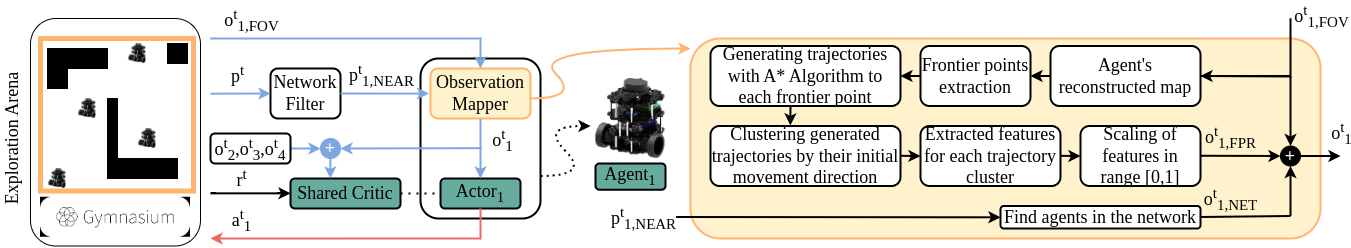}
\caption{Illustration of the centralized training framework for collaborative multi-agent exploration in unknown environments. The scenario involves four robots navigating an occupancy grid, where black cells denote obstacles and white cells indicate free space, implemented using Gymnasium. The diagram focuses on Agent 1, which gathers raw data from the arena and processes them through the observation mapper to compute the observation used by the actor policy to select the next action. Implementation details of this module are shown in the yellow section. Additionally, the shared critic, which receives observations from all agents along with the joint reward, facilitates actor policy training. The other agents follow a similar architecture to Agent 1.}
\label{fig:concept}
\end{figure*}

\section{Related Work}
Extensive research on decentralized collaborative reinforcement learning (RL) architectures for multi-agent exploration has been proposed over the years \cite{da2020agents,zhang2021multi,liu2022review,yadav2023comprehensive,hu2024review, ning2024survey} and has focused on various key aspects, such as RL algorithms, challenges, and applications. The approach proposed in \cite{li2024efficient} includes an Efficient Multi-Agent Cooperation (EMAC) approach, which integrates a Heterogeneous Graph Network (HGN) for extracting semantic information from local observations and a Frame-wise Communication Reduction Algorithm (FCRA) approximating multi-hop communication using 1-hop communication through historical embeddings for scalable multi-agent reinforcement learning (MARL) under partial observability and communication constraints. \cite{cao2023ariadne} proposes ARiADNE, an attention-based neural approach for real-time, non-myopic path planning in autonomous exploration, which learns dependencies across multiple spatial scales in the agent's partial map and predicts potential gains from exploring these areas, enabling the agent to effectively balance between known regions and exploring new ones. 

Moreover, \cite{psomiadis2024multi} addresses the challenge of exploring unknown environments with mobile sensors that communicate compressed measurements, introducing a novel communication framework paired with a tractable multi-agent exploration algorithm designed to optimize sensor actions, utilizing a task-driven measure of compression uncertainty as the reward function. Notably, a multi-robot environmental exploration strategy called Hierarchical-Hops Graph Neural Networks (H2GNN) leveraging multi-hop information to enable attentive environmental perception and incorporating multi-agent reinforcement learning (MARL) for autonomous strategy learning is proposed in \cite{zhang2022h2gnn} and demonstrate to foster collaborative behavior and improve exploration efficiency. Instead, \cite{xue2023multi} proposes a novel Multi-Agent Recurrent Deterministic Policy Gradient (MARDPG) algorithm, based on the Depth Deterministic Policy Gradient (DDPG) framework, for multi-UAV navigation control, where each UAV’s critic network is centrally trained to estimate other UAVs' policies while the decentralized execution eliminates the need for inter-UAV communication and the LSTM network enhances action-value prediction by using historical exploration data. 

In \cite{yan2023mui}, Multi-agent TARE with Unknown Initial Position (MUI-TARE), a multi-agent lidar-based exploration system integrating adaptive merging based on AutoMerge and a hierarchical cooperative multi-agent planning strategy for exploration manages to optimize the balance between sub-map merging robustness and exploration efficiency. Moreover, a theoretical framework called Cyber-Physical Partially Observable Markov Decision Process (CP-POMDP) to jointly optimize control and communication through MARL is proposed in \cite{mason2024multi}, which integrates two types of agents that perform actions affecting the physical environments, and sensors to communicate their observations to the robots through resource-constrained communication channels. Remarkably, \cite{zhang2021centralized} investigates a MARL architecture with Centralized Models and Exploration (MARCO) that is a model-based reinforcement learning method leveraging a centralized stationary model independent of the agents' policies for sample-efficient learning in Dec-POMDPs. It employs a model-free MARL algorithm for policy optimization and a centralized exploration policy to enhance sample complexity by targeting high-uncertainty regions of the state-action space, and uses R-MAX, a model-based dynamic programming algorithm, for Markov Decision Processes for tabular Dec-POMDPs.

\section{Methodology}
This section provides an overview of the environment that needs to be explored, as described in \ref{sbs:overview}, introduces the Dec-POMDP architecture used to model the problem in \ref{sbs:dec_pomdp}, and discusses the HAPPO algorithm in \ref{sbs:happo} employed to optimize the agents' policies and shared critic network in \ref{sbs:policies}. Finally, the formulation of the reward functions is extensively discussed in \ref{sbs:rewards}.

\subsection{Overview of the RL environment}
\label{sbs:overview}
The agents are required to collaboratively map a flat exploration arena with obstacles that is represented by an occupancy grid $\mathcal{G}^t$ at time $t$ of size $m \times n$, such that the following condition holds:
\begin{equation}
\label{eq:definition_g}
    \mathcal{G}_{kw}^t = 
    \begin{cases}
    0, & \text{if the grid } (k, w) \text{ is free}, \\
    1, & \text{if the grid } (k, w) \text{ is occupied}.
    \end{cases}
\end{equation}
It is assumed that an occupied grid $\mathcal{G}_{kw}^t$ can either be filled by a static obstacle or by a dynamic obstacle, i.e., an agent located in the cell. Furthermore, each grid is assumed to be square-shaped with a side length denoted by $\Delta$. The value of this parameter is considered as a trade-off between a sufficiently detailed representation of the operating space, which favors finer values for $\Delta$, and limited memory capacity for the agent's map storage, which benefits from larger $\Delta$.

\subsection{Decentralized-POMDP architecture}
\label{sbs:dec_pomdp}
The decision-making scenario for multi-agent exploration has been modeled through a Dec-POMDP \cite{oliehoek2016concise}, a specialization of a Partially Observable Stochastic Game \cite{marl-book}, synthesized by the tuple $\langle \mathbb{I}, \mathbb{S}, \mathbb{A}, T, \mathbb{O}, O, R, h, b_0 \rangle$. At time $t$, each available exploring agent $i \in \mathbb{I}$ acquires an observation $o_i^t \in \mathbb{O}$ partially describing the environment's state $s^t \in \mathbb{S}$, and chooses an action $a_i^t \in \mathbb{A}$ according to the probability distributions $O$ and $T$, respectively. The other terms are the immediate reward $R$, the horizon of the problem $h$, and the initial state distribution $b_0$. In particular, the environment's state $s^t$ includes the agents' positions $p^t = (p_i^t)_{i \in \mathbb{I}}$, and the updated occupancy grid $\mathcal{G}^t$, while the action space $\mathbb{A}$ is designed such that each agent $i \in \mathbb{I}$ can choose between 10 actions which include the 8 movements to the adjacent grid cells, an action to remain in the current position, and a communication action to share exploration data with other agents within the same data-transmission network $\mathcal{N}$. 

As shown in Fig. \ref{fig:concept}, each agent $i$ locally observes the environment through a squared agent-centered FOV occupancy grid $o_{i,FOV}^t = \{\mathcal{G}_{kw}^t \in \mathcal{G}^t \mid \| (k, w) - p_i^t \|_2 \leq r_d \}$, where $p_i^t \in \mathcal{G}^t$ and $r_d$ are the position and the sensing range of agent $i$, respectively, and $\|\cdot\|_2$ denotes the Euclidean norm. The FOV map is also used to update the agent's reconstructed map of the environment, expressed as $M_i^t = M_i^{t-1} \cup o_{i,FOV}^t$, as illustrated in Fig. \ref{fig:concept}. Specifically, \( o_{i,FOV}^t \) is used to identify obstacle regions in the agent's surroundings, while the reconstructed map is leveraged to detect frontier points. As a result, \( o_{i,FOV}^t \) marks cells with dynamic obstacles as occupied, while the reconstructed map classifies these same cells as free. To address the challenge of increasing size of the map $M_i^t$ during exploration, we propose a pre-processing stage that extracts a feature table $o_{i,FPR}^t$ indicating Frontier Points Reachability (FPR) according to Algorithm \ref{alg:fpr}. This input to the agent's policy enables the robot to prioritize movements toward regions in the environment with the highest density of frontier points, while also taking the estimated navigation distances into consideration.

\begin{algorithm}
\caption{Feature extraction from frontier trajectories}
\label{alg:fpr}
\begin{algorithmic}
\Require Reconstructed environment $M_i^t$, agent position $p_i^t$
\Ensure Feature table $o_{i,FPR}^t$

\State $F_i^t (M_i^t) \gets $ the frontiers $ \{ m_i^t \in M_i^t \mid m_i^t \in \text{Free}(M_i^t) \wedge \exists m_i^{t'} \in \text{Unexplored}(M_i^t), \, \|m_i^t - m_i^{t'}\|_1 = 1 \} $
\State \( T_i^t \gets\) the trajectories from \( p_i^t \) to each frontier \( f_i^t \in F_i^t\) using A* algorithm \cite{foead2021systematic}
\For{each direction $d$ to an adjacent cell of $p_i^t$}
    \State $T_{i,d}^t \gets \{ T_i^t \mid \text{FirstMove}(T_i^t) = d \}$, i.e. the trajectories leading to the frontiers whose first move corresponds to $d$
    \State $n_d \gets $ the number of trajectories $ \mid T_{i,d}^t \mid$
    \State Extract the average $\mu_d$ and standard deviation $\sigma_d$ of the estimated trajectory lengths.:
    \State $\mu_d \gets \frac{1}{N_d} \sum_{T \in T_{i,d}^t} \text{Length}(T)$
    \State $\sigma_d \gets \sqrt{\frac{1}{N_d} \sum_{T \in T_{i,d}^t} (\text{Length}(T) - \mu_d)^2}$
    \State Add $n_d$, $\mu_d$ and $\sigma_d$ to $o_{i,FPR}^t$
\EndFor
\State Normalize each feature in $o_{i,FPR}^t$ to the range [0,1]

\end{algorithmic}
\end{algorithm}
In particular, the observation $o_{i,FPR}^t$ provides the policy with a fixed-length input capturing the frontier points' distribution, independently of the extension of the exploration arena. Furthermore, Fig. \ref{fig:concept} shows that each agent detects the positions of the other robots that are within its communication-covered zone which can be expressed as $p^t_{i,NEAR} = \{p_k^t \in p^t \mid \| p_k^t - p_i^t \|_2 \leq r_c \}$ where $r_c$ is the agent i's communication range. Therefore, the observation mapper uses this information to compute the vector $o_{i,NET}^t$ which indicates whether agent $i$ can communicate with each exploring agent. Therefore, at each time $t$, the agent $i$'s policy receives an observation $o_i^t = \left( o_{i,FOV}^t, \, o_{i,FPR}^t, \, o_{i,NET}^t \right)$.

\paragraph{Inter-agent communication strategy}
\label{prg:communication}
The implemented strategy to account for inter-agent data transmission involves two-way explicit communication where if an agent decides to share its reconstructed map $M_i^t$, it broadcasts to all the agents within the same transmission network $\mathcal{N} \subseteq \mathbb{I}$. Two conditions must be satisfied for a pair of agents to be considered as belonging to the same network: (1) only agents that have explicitly chosen to communicate can be part of the network, and (2) agents must be in close proximity. The latter condition is satisfied if agents \(i\) and \(j\) are within each other's communication-covered zone or if there exists a chain of intermediate agents such that each pair meets this distance criterion. Communicating agents in the same network share the reconstructed maps, and merge them into a single map $M_\mathcal{N}^t = \bigcup_{j \in \mathcal{N}} \mathcal{M}_j^t$. This shared map thus replaces the reconstructed maps of each agent in the network.

\begin{figure*}[t]
\centering
\includegraphics[width=0.9\linewidth]{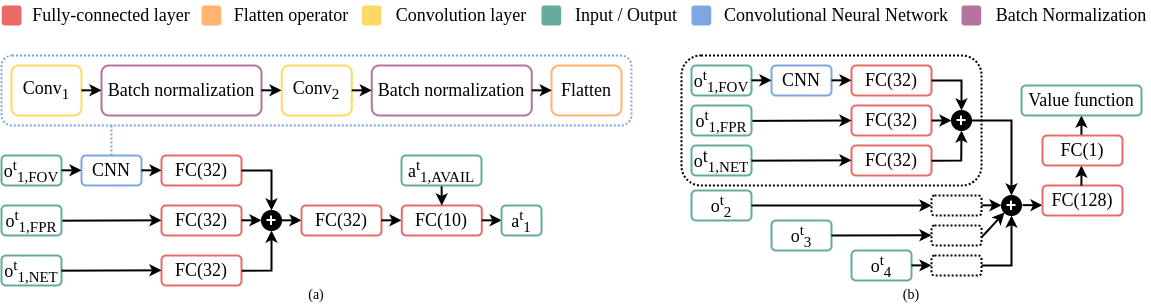}
\caption{Neural network architectures based on convolutional layers for the modeling of the agent policy (a) and shared critic (b) in the multi-agent exploration framework. In particular, while the left figure illustrates the architecture of agent $1's$ policy $\pi_1$, analogous structures are valid also for the other agents.}
\label{fig:policy}
\end{figure*}

\subsection{Multi-agent reinforcement learning algorithm}
\label{sbs:happo}
The employed framework follows the Centralized Training with Decentralized Execution (CTDE) paradigm, where the shared critic has access to the agents' joint observations, $o^t = (o_i^t)_{i \in \mathbb{I}}$, and the collective rewards $r^t = (r_i^t)_{i \in \mathbb{I}}$ during training, as depicted in Fig. \ref{fig:concept}. However, during execution, the critic is no longer present, and each agent operates independently, relying solely on its local observations without access to any centralized data. The Heterogeneous-Agent Proximal Policy Optimization (HAPPO) algorithm \cite{zhong2024heterogeneous} is employed for optimizing the policies of the exploring robots. Each agent $i$ is equipped with an agent-specific deep neural network policy $\pi_i(\theta_i)$, which determines the action $a_i^t$ based on the observation $o_i^t$. To support policy evaluation and improvement, a shared critic network $V(\phi)$ is used to estimate the value function. In particular, HAPPO sequentially updates during training each agent’s policy through the PPO pipeline, adjusting the parameters $\theta_i^t$ of the policy $\pi_i$ by leveraging the advantage function, which is computed from both the current trajectory and previously updated advantage values. Instead, the parameters $\phi^t$ of the critic network $V$ are updated by minimizing the mean squared error between the predicted state value nad the target return.

\subsection{Design of the actors and shared critic architectures}
\label{sbs:policies}
Two neural architectures for modeling the joint actor' policy $\{\pi_i(\theta_i)\}_{i \in \mathbb{I}}$ and the shared critic network $V(\phi)$ have been analyzed and are described in detail below.
\begin{itemize}
    \item \textbf{Multilayer Perceptron (MLP)}. This architecture takes as input the concatenation of $o^t_{i,FOV}$, the flattened $o^t_{i,FPR}$ and $o^t_{i,NET}$ and extract features by using two fully connected layers, separated by layer normalization, described in Table \ref{tab:first_policy}. Then a final fully connected layer predicts the action $a_i^t$ by using a categorical distribution and masking the actions that are not feasible via the input \(a_{i,AVAIL}^t\) as a safety guarantee for real-world applications. The shared critic employs an analogous architecture but concatenates the components of the joint observation \(o^t\) as input, resulting in the layer \(Linear_1\) having 148 input features. In this case, the final layer does not include a categorical layer and returns the value function estimate as a scalar value. In both cases, the dimensions of the input features for $Linear_1$ depend on the environment parameters in Table \ref{tab:params}.
    \begin{table}
    \centering
    \caption{Hyperparameters for the layers in the MLP architecture}
    \label{tab:first_policy}
    \begin{tabular}{lcccc}
        \toprule
        \textbf{Layer} & \textbf{Input features} & \textbf{Output features} & \textbf{Bias} & \textbf{Activation}\\
        \midrule
        $Linear_1$ & 37 & 2400 & Yes & ReLU\\
        \midrule
        $Linear_2$ & 2400 & 300 & Yes & ReLU\\
        \midrule
        $Linear_3$ & 300 & 10 & Yes & None\\
        \bottomrule
    \end{tabular}
    \end{table}
    \item \textbf{Convolutional Neural Network (CNN)}. This architecture is illustrated in detail in Fig. \ref{fig:policy} for both the actor's policy and the shared critic. In particular, a CNN, with hyperparameters provided in Table \ref{tab:conv_layers}, is employed to extract spatial features from the agent-centered FOV occupancy grid \(o_{i,FOV}^t\). The CNN output is processed by a fully connected layer and combined with the features from the fully connected layers applied on \(o_{i,FPR}^t\) and \(o_{i,NET}^t\). Then, a multilayer perceptron selects the action \(a_i^t\), with the final layer similar to the previous architecture. The shared critic uses a similar design, initially processing each agent's observations separately before concatenating them and passing through an MLP to compute the value function estimate.
    
    \begin{table}[h]
    \centering
    \caption{Hyperparameters for the convolution layers for the CNN architecture}
    \label{tab:conv_layers}
    \begin{tabular}{lcccc}
        \toprule
        \textbf{Layer} & \textbf{Input channels} & \textbf{Output channels} & \textbf{Kernel} & \textbf{Stride} \\
        \midrule
        $Conv_1$ & 1 & 8 & 3 & 1\\
        \midrule
        $Conv_2$ & 8 & 16 & 3 & 1\\
        \bottomrule
    \end{tabular}
\end{table}
\end{itemize}

\begin{figure*}[h]
	\centering
	\begin{subfigure}{0.50\linewidth}
		\includegraphics[width=\linewidth]{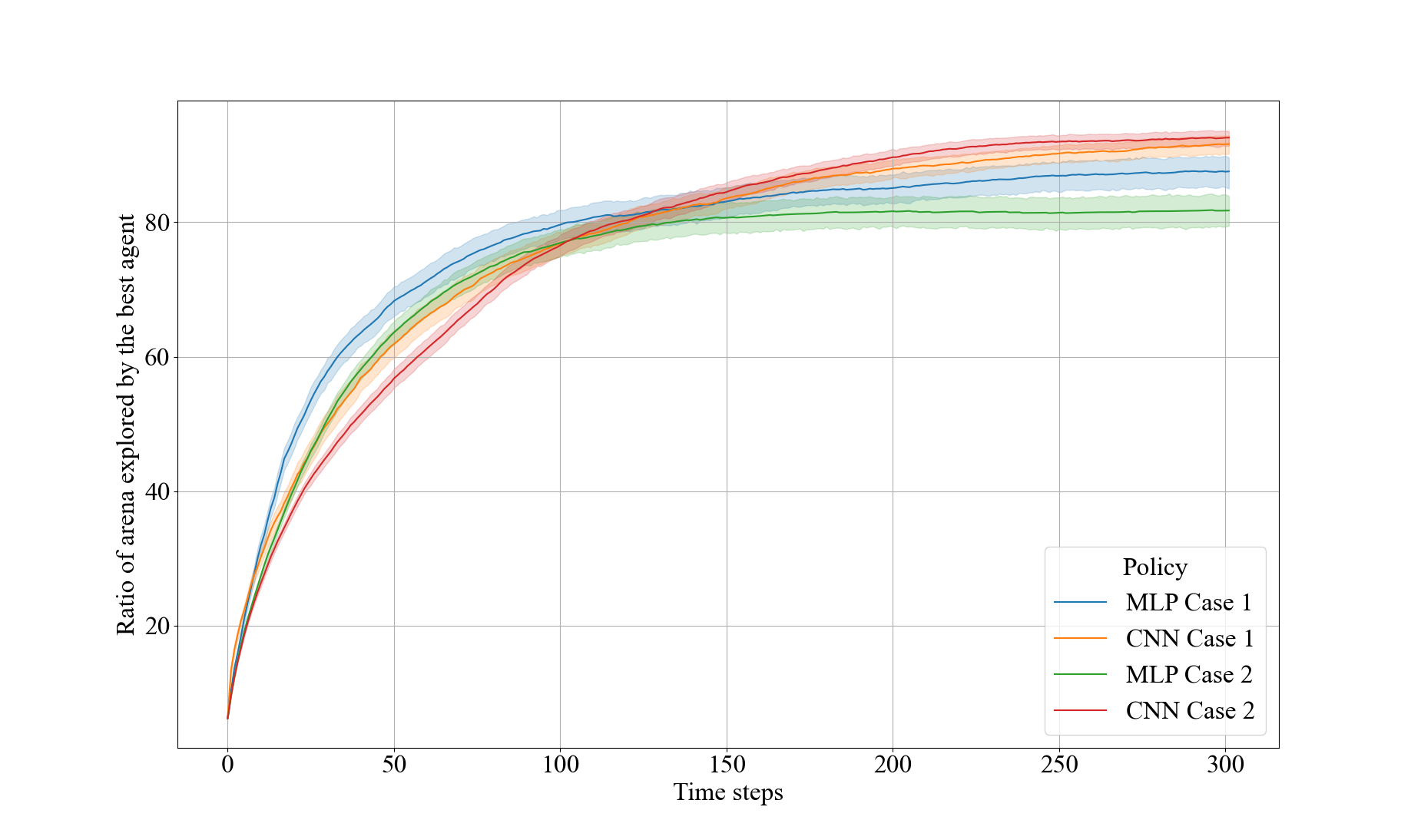}
	  \caption{Exploration ratio of the environment by the agent with the largest global map at each time across all simulations.}
		\label{fig:simulation_of_exp}
	\end{subfigure}%
	\begin{subfigure}{0.50\linewidth}
		\includegraphics[width=\linewidth]{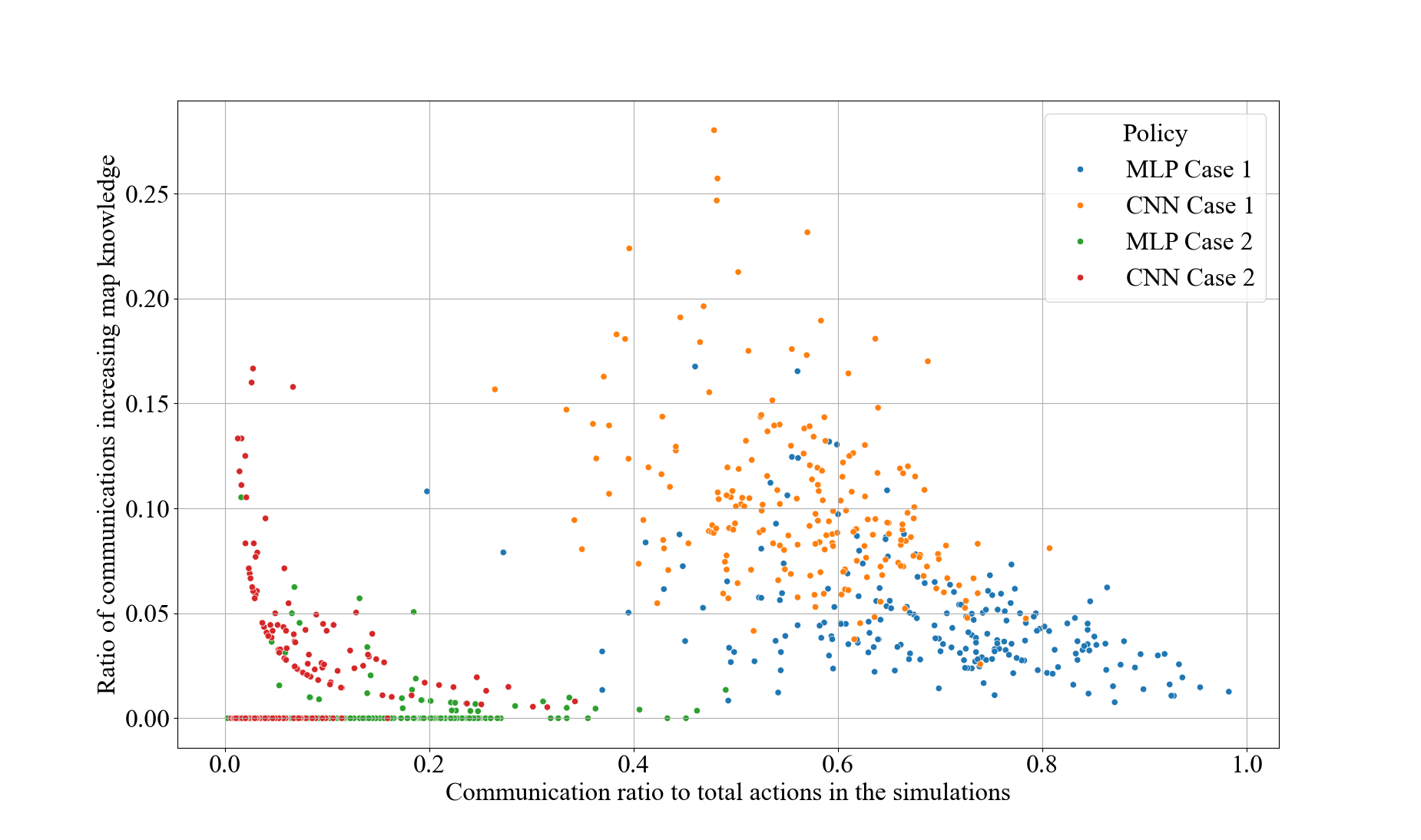}
		\caption{Illustration of the ratio of communication actions to total actions for each simulation, along with the ratio of communication events that expanded the agents' map over all data transmissions.}
		\label{fig:successful_communication}
	\end{subfigure}%
   
	\caption{Statistics from simulating the trained policies, evaluated across 200 exploration arenas with randomized static obstacles and agent initial positions, using the two proposed neural network architectures and reward functions.}
	\label{fig:sim_graphs}
\end{figure*}

\begin{figure}[b]
\centering
\includegraphics[width=\linewidth]{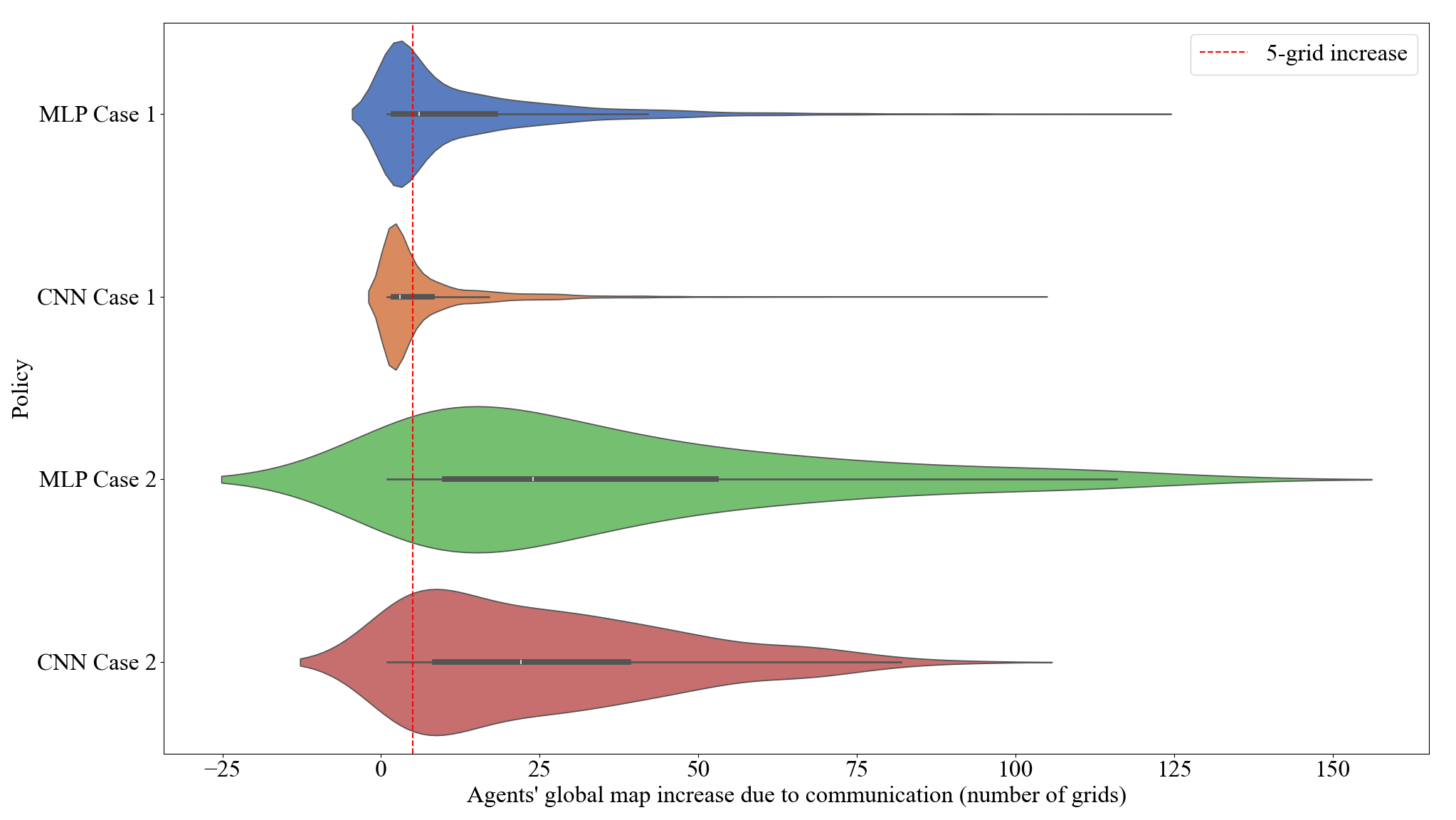}
\caption{Distribution of agents' map expansions due to inter-agent communication, for the four policy types across the 200 simulated exploration arenas.}
\label{fig:communication_reward}
\end{figure}

\subsection{Communication-induced reward function}
\label{sbs:rewards}
Since the multi-agent decision-making problem investigated in this research is modeled as a Dec-POMDP, all agents have the same rewards. In particular, two reward functions have been considered to train agents in exploratory and communicative behaviors for effective exploration and the implementation details are described below.
\begin{itemize}
    \item \textbf{Case 1}: This reward function formalized in Eq. \eqref{eq:reward_1} accounts for both the safety of the action $a_i^t$ selected by agent $i$ and the number of unknown grids discovered in the exploration arena due to the action itself. In particular, agents that perform dangerous actions, such as colliding with the arena boundaries or obstacles, receive a penalty of -100. Otherwise, if the agent shares its reconstructed map with other agents or it explores actively the environment, it is rewarded based on the number of cells added to its map. Therefore, communication is favored over exploration whenever it leads to a greater number of cells being acquired.
    \begin{equation}
    \label{eq:reward_1}
    r_i^t = 
    \begin{cases} 
    -100 & \text{if } a_i^t \text{ is dangerous action,} \\
    M_\mathcal{N}^t - M_i^{t-1} & \text{if } i \text{ shares data in } \mathcal{N} \text{,}\\
    M_i^t - M_i^{t-1} & \text{if } i \text{ actively explores.}
    \end{cases}
    \end{equation}
    
    \item \textbf{Case 2}: This reward function has been proposed in \cite{marl_exp} and its structure is depicted in Eq. \eqref{eq:reward_2}.
    \begin{equation}
    \label{eq:reward_2}
    r_i^t = 
    \begin{cases} 
    -10 & \text{if } a_i^t \text{ is dangerous action,} \\
     p_i^t \frac{M_\mathcal{N}^t - M_i^{t-1}}{m \times n}  & \text{if } i \text{ shares data in } \mathcal{N} \text{,}\\
    \frac{M_i^t - M_i^{t-1}}{e_{max}} - r^t_{i,REP} & \text{if } i \text{ actively explores.}
    \end{cases}
    \end{equation}
    where 
\[
p_i^t = \frac{1}{\left| \mathcal{N} \right| - 1} \sum_{j \in \mathcal{N}, j \neq i} \frac{q_{ij}}{m \times n} + 0.8
\]
In particular, \( q_{ij} \) denotes the number of discoveries made by agent \( i \) since its last communication with agent \( j \) and $e_{max}$, as defined in Eq. \eqref{eq:e_max}, represent the maximum potential increase in environmental knowledge per step, where $r_d$ is the agent's detection range and $\Delta$ is the exploration arena's grid dimension. 

\begin{equation}
    \label{eq:e_max}
    e_{max} = 2 \left(2 \frac{r_d}{\Delta} + 1 \right) - 1 = 4 \frac{r_d}{\Delta} + 1
\end{equation}

Additionally, $r^t_{i,REP}$ is -1 if the agent is stationary and 0 otherwise, to discourage idleness unless the agent is sharing data.
\end{itemize}

In both cases, the final rewards for each agent are computed as the average of the individually collected rewards.

\section{Training and simulations}
The training and policy evaluations have been executed on a computer with Ubuntu 22.04.4 LTS equipped with 32 14th Generation Intel® Core™ i9-14900K and an NVIDIA GeForce RTX 4090 with 24,564 MiB. Gymnasium \cite{towers2024gymnasium} and the PettingZoo library \cite{terry2021pettingzoo} with the Parallel API have been employed to implement the multi-agent reinforcement learning problem, while PyTorch is designed for the neural network implementation. 
Additionally, the choice of the parameters for defining the exploration arena for policy training, as shown in Table \ref{tab:params}, adheres to the experiment setup explained in \ref{sct:experiment} to ensure consistency with the real-world simulations. In particular, the policies have been trained in environments with varying obstacles and initial agents' positions, maintaining a constant ratio of static obstacles to the total number of cells at $\rho_{obstacle}$, to enhance the agents' adaptability to various scenarios. Moreover, Table \ref{tab:params} reports the key hyper-parameters for training. Indeed, $n_e$, $n_s$, $n_b$, $\epsilon$, $ppo_{epoch}$ indicate the number of episodes, the steps per episode, the batch size used for updating the actors and shared critic, the clipping range limiting the policy update, the number of epochs for which the PPO algorithm optimizes the collected data, respectively. Furthermore, the learning rate for the actor and critic networks is set to $5 \times 10^{-4}$.

\begin{table}[h]
    \centering
    \caption{Parameters used for environment and training setup}
    \label{tab:params}
    \begin{tabular}{lccccc}
        \toprule
        \multicolumn{6}{c}{\textbf{Environment}} \\
        \midrule
        \textbf{Name} & $n, m$ & $\rho_{obstacle}$ & $\Delta$ & $r_d$ & $r_c$ \\
        \midrule
        \textbf{Value} & 12 & 0.1 & 0.5 & 1.1 & 3.2 \\
        \toprule
        \multicolumn{6}{c}{\textbf{Training}} \\
        \midrule
        \textbf{Name} & $n_e$ & $n_s$ & $n_b$ & $\epsilon$ & $ppo_{epoch}$\\
        \midrule
        \textbf{Value} & 5,000 & 200 & 8 & 0.2 & 5\\
        \bottomrule
    \end{tabular}
\end{table}

\begin{figure*}[h]
\centering
\includegraphics[width=\linewidth]{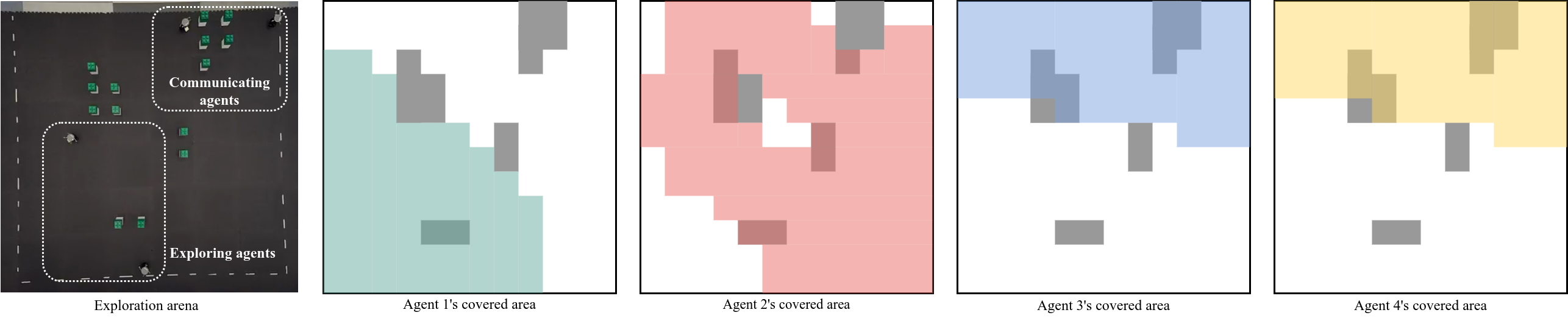}
\caption{From left to right: the first image depicts the custom-built lab setup with four TurtleBot3 mobile platforms, two communicating and two exploring, in the exploration arena outlined by the dashed line. In particular, the obstacles are indicated by the green blocks. The remaining four figures on the right show the areas explored by each agent, represented by different colors, while the static obstacles are highlighted in gray.}
\label{fig:concept_lab}
\end{figure*}

\begin{figure}[b]
\centering
\includegraphics[width=\linewidth]{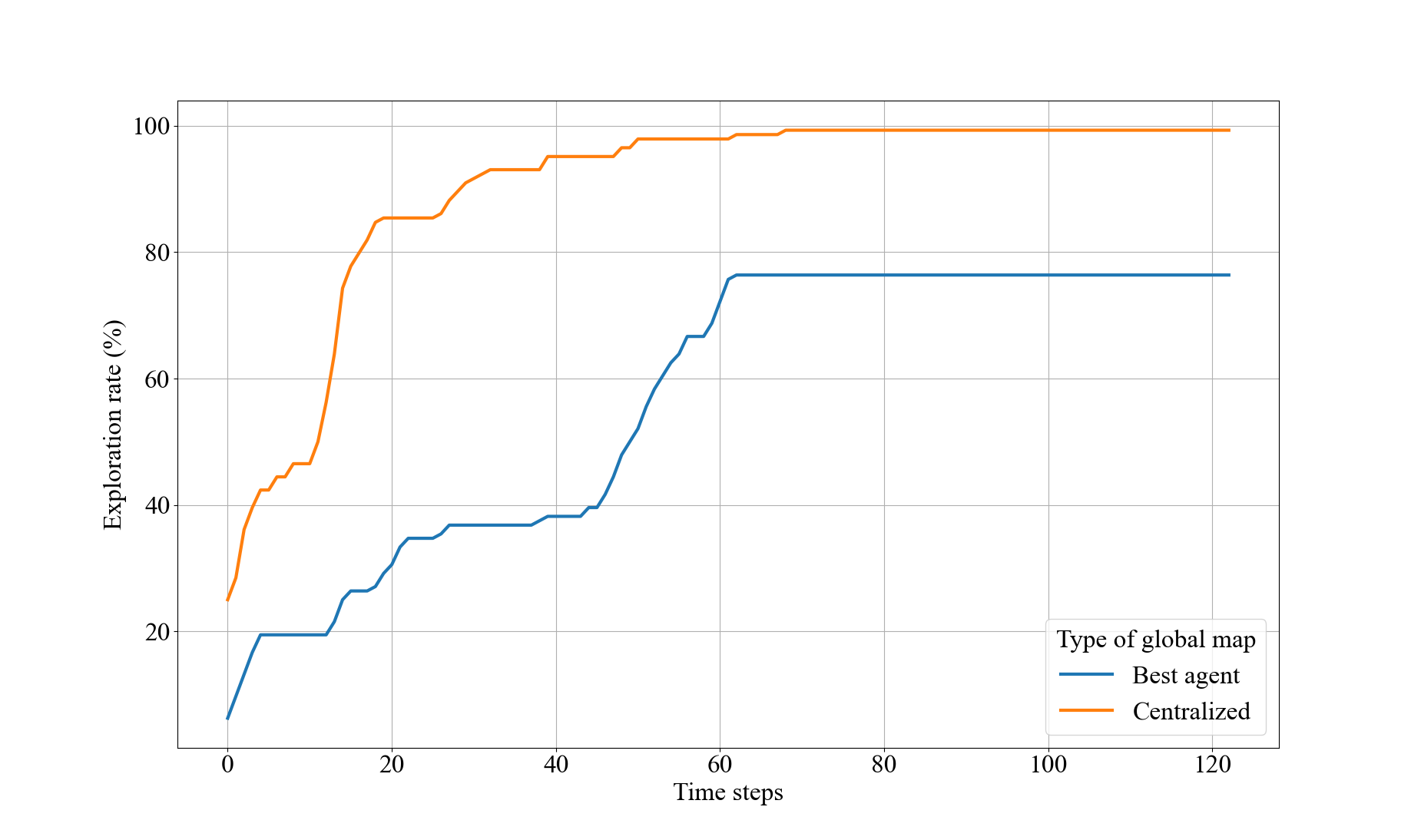}
\caption{Exploration rates of the robots in the custom-built exploration arena in the lab environment, based on the agent with the largest global map and the map obtained by merging all agents' global maps at each time step of the experiment.}
\label{fig:exploration_rate_experiment}
\end{figure}

\subsection{Simulation results}

The trained policies are tested in 200 exploration arenas with random static obstacles and initial agents' positions to evaluate the multi-agent exploration framework's effectiveness across different environments. Thus, each set of agents, trained using either MLP or CNN architectures and reward functions Case 1 or Case 2, is assessed over 300 time steps in these new environments. 

Figure \ref{fig:simulation_of_exp} illustrates the exploration progress of the agent with the largest map at each time step for all the simulations. It shows that all policies maintain similar exploration rates independently on the configuration of the exploration arena, with the collected data closely centered around the average. Specifically, all types of agents achieve at least 80\% exploration after 150 time steps, and CNN-based policies surpass 90\% by the end. Remarkably, the best performance is observed with policies trained with the reward function Case 2 with exploration rates reaching up to 99\%. 

Figure \ref{fig:successful_communication} depicts the proportion of communication actions relative to the total number of selected actions and their effectiveness in expanding the global map. It reveals that reward function Case 1 leads to a higher frequency of communication actions over exploration actions, often resulting in stationary agents. However, these communications are generally more effective with CNN-based policies. Conversely, agents trained with reward function Case 1 show reduced communication in favor of exploration, with CNN-based policies demonstrating superior communication efficiency over MLP-based ones. This different balance between communication and exploration between Case 1 and 2 is also evident in Fig. \ref{fig:simulation_of_exp}, where agents trained with the reward in Case 1 initially have larger maps due to higher communication rates, but policies developed according to Case 2 ultimately perform better by maintaining consistent exploration. Moreover, Fig. \ref{fig:communication_reward} displays the distribution of map expansions due to inter-agent communication across the four policy types. In particular, when the expansion is greater than 5 cells, which is the maximum number of cells that can be explored following a diagonal movement, i.e., $e_{max}$, opting for communication rather than exploration yields greater benefits. Figure \ref{fig:communication_reward} shows that agents trained with reward function Case 2 share data with the greater agents' advantage than the other policies since the distributions show a more pronounced rightward skew.

\section{Experimental validation}
\label{sct:experiment}

To evaluate the trained agents for multi-agent exploration, a framework based on Robot Operating System 2 (ROS2) Humble Hawksbill has been designed to replicate the simulations in a custom-built exploration arena in lab environment with four Turtlebot3 mobile robots. As depicted in Fig. \ref{fig:concept_lab}, the exploration arena consists of a 6m $\times$ 6m space, outlined by white dashed lines, with static obstacles constructed from green blocks positioned within the occupied grid cells. In particular, the environment parameters used in the experiment are consistent with the ones mentioned in Table \ref{tab:params}. 
The implemented framework associates to each robot $i$ a ROS2 node responsible for collecting the raw data from the environment, computing the observation $o_i^t$ through the observation mapper whose implementation details are shown in Fig. $\ref{fig:concept}$, processing the action $a_i^t$ through the policy $\pi_i$ and applying it to the environment. To ensure proper coordination between the robotic platforms, a shared clock is initialized simultaneously at the beginning of the experiment. Regarding the data collection process, each agent gathers a 360° laser scan describing its surroundings during every execution cycle using the 2D LDS-01 Laser Distance Sensor mounted on the robotic platform. Additionally, the agent’s position and orientation in the global reference frame associated with the exploration arena are provided by the Vicon motion capture system. Each movement action selected by a robot is executed by a navigation controller which is responsible for guiding the robot to the center of the neighboring cell indicated by the selected action. In the case of diagonal movements, the robot sequentially navigates through two adjacent cells to reach the final goal. On the other hand, communication is implemented by sharing through ROS2 topics the global map acquired by the robots. For instance, Fig. \ref{fig:concept_lab} shows that agent 3 and 4 have similar coverages. That is due to a communication event close to the end of the experiment where the two agents merged their maps.

Figure \ref{fig:exploration_rate_experiment} shows the exploration rate of the agent with the largest global map at each time step and the merged map from all agents. The results shows that after 60 steps, the best agent covers around 80\% of the environment, as it is depicted also in agent 2's map in Fig. \ref{fig:concept_lab}. Moreover, Fig. \ref{fig:exploration_rate_experiment} illustrates that the merged map steadily grows reaching nearly 100\% exploration by 60 time steps. 

\section{Conclusions}
This study presents a decentralized collaborative framework for multi-agent exploration in unknown, obstacle-filled environments using reinforcement learning. The proposed architecture leverages sensor data easily available from real robotic platforms and introduces a novel methodology to estimate the optimal navigation directions based on frontiers distribution. The results show that inter-agent proximity-based communication significantly reduces exploration time by enabling agents to share their global maps and minimize redundant efforts. Moreover, experimental validation in a real-world exploration arena demonstrates the framework's feasibility with real robots. Future work will focus on extending this framework to heterogeneous robotic platforms to further enhance multi-agent exploration.

\bibliographystyle{IEEEtran}
\bibliography{main}

\end{document}